\documentclass[10pt,twocolumn,letterpaper]{article}

\usepackage{iccv}
\usepackage{times}
\usepackage{epsfig}
\usepackage{graphicx}
\usepackage{amsmath}
\usepackage{amssymb}

\usepackage{subcaption}
\usepackage[dvipsnames]{xcolor}
\usepackage[normalem]{ulem}

\usepackage{authblk}
\makeatletter
\renewcommand\AB@affilsepx{\\  \protect\Affilfont}

\makeatother

\usepackage[pagebackref=true,breaklinks=true,letterpaper=true,colorlinks,bookmarks=false]{hyperref}

\iccvfinalcopy 

\def\iccvPaperID{****} 

\begin{document}

\title{Deep Camera Obscura: An Image Restoration Pipeline for \\Lensless Pinhole Photography}

\author[1]{Joshua D. Rego}
\author[2]{Huaijin Chen}
\author[2]{Shuai Li}
\author[2]{Jinwei Gu}
\author[1]{Suren Jayasuriya}

\affil[1]{Arizona State University, Tempe, AZ}
\affil[2]{SenseBrain Technology, San Jose, CA}

\affil[1]{{\tt\small \{jdrego, sjayasur\}@asu.edu} \qquad  
         $^2${\tt\small \{chenhuaijin, shuai.li, gujinwei\}@sensebrain.ai}}

\twocolumn[{%
\renewcommand\twocolumn[1][]{#1}%
\maketitle
\begin{center}
    \centering
    \captionsetup{type=figure}
    \includegraphics[width=\textwidth]{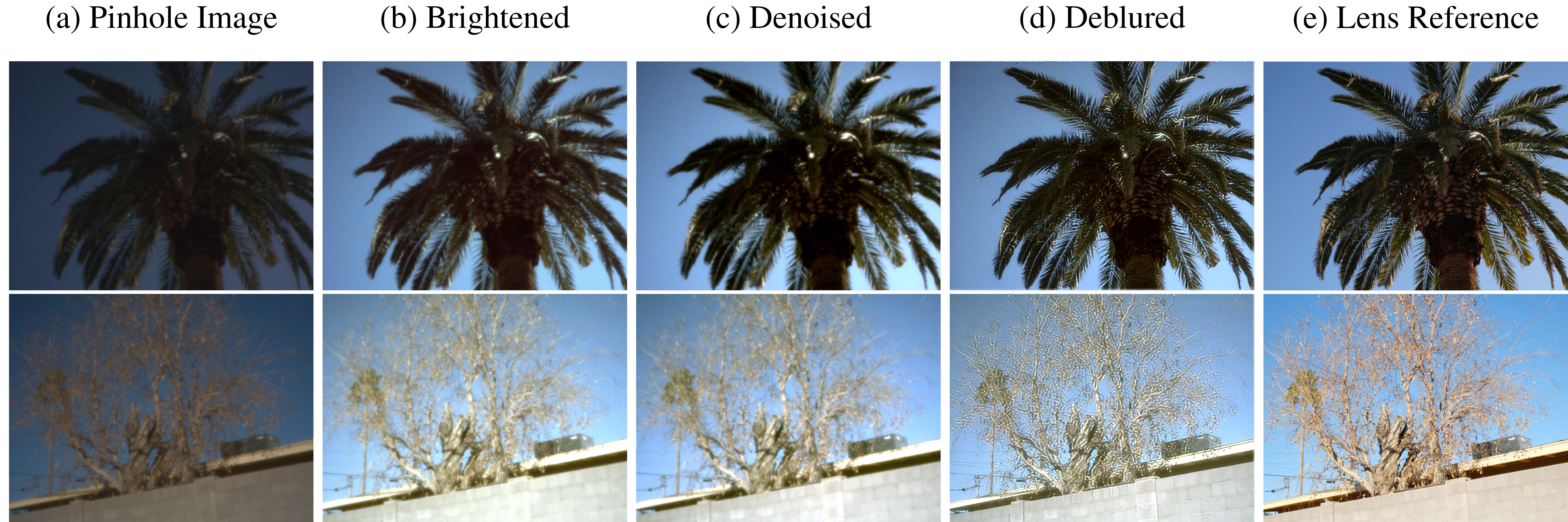}
    \caption[PipelineResults]{The proposed Deep Camera Obscura (DCO) pipeline is a jointly optimized denoise + deblur pipeline that restores degraded lensless pinhole camera images suitable for perceptual viewing. Our pipeline is trained on synthetic data and utilizes domain knowledge of the optical point spread function to help improve image restoration for pinhole cameras. The resulting DCO pipeline can operate on 20 MP images with 1/30s exposure time.}
    \label{fig:palm-result}
\end{center}
}]

\begin{abstract}
    The lensless pinhole camera is perhaps the earliest and simplest form of an imaging system using only a pinhole-sized aperture in place of a lens. They can capture an infinite depth-of-field and offer greater freedom from optical distortion over their lens-based counterparts. However, the inherent limitations of a pinhole system result in lower sharpness from blur caused by optical diffraction and higher noise levels due to low light throughput of the small aperture, requiring very long exposure times to capture well-exposed images. In this paper, we explore an image restoration pipeline using deep learning and domain-knowledge of the pinhole system to enhance the pinhole image quality through a joint denoise and deblur approach. Our approach allows for more practical exposure times for hand-held photography and provides higher image quality, making it more suitable for daily photography compared to other lensless cameras while keeping size and cost low. This opens up the potential of pinhole cameras to be used in smaller devices, such as smartphones.
\end{abstract}

\section{Introduction}
\label{sec:intro}

    \begin{figure*}
        \centering
        \includegraphics[width=\textwidth]{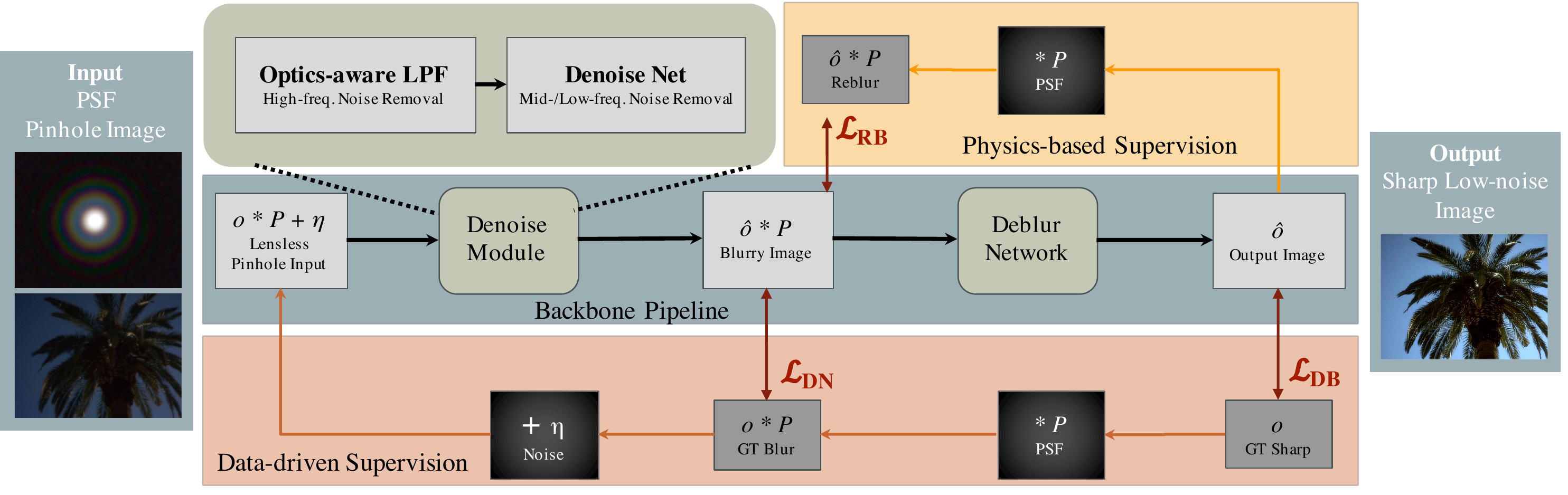}
        \vspace{-5mm}
        \caption{\textbf{System Architecture:} Due to diffraction and low light throughput, pinhole camera images observe two major artifacts: optical blur and sensor noise, for which we propose a jointly optimized denoise and deblur framework to tackle.
        The proposed system turns blur into an advantage for better denoising. Knowing a circular pinhole automatically performs ideal low-pass filtering (LPF) in optics (see sec. \ref{sec:approach}), the denoise module first performs optics-aware LPF with an ideal LPF that matches the pinhole's diffraction limit, since any signal with a frequency above the limit is due to noise. A denoise network then further cleans mid- and low-frequency noise, given deblur or deconvolution process is commonly prone to noise. Finally, a GAN-based deblur network is used to recover high-frequency information lost in the pinhole imaging process to form the final output. Both modules are jointly trained with data-driven loss and physical consistency to eliminate GAN-style artifacts.}
        \vspace{-4mm}
        \label{fig:pinhole-pipeline}
    \end{figure*}
    
    Cameras today have come a long way in enabling high-quality photography capabilities to be easily accessible. Smartphones now incorporate multiple cameras with sensors that can exceed 50MP and enable pro-level features such as HDR mode, panorama stitching, and augmented reality. While these advancements have greatly improved the creativity and quality of everyday photography, lens-based cameras can have their own limitations. Traditionally, smartphone image quality has been largely limited by the image sensor size that can be manufactured. Larger sensors allow for higher quality images in terms of resolution, noise, and low-light detection. However, larger sensors also require larger, bulkier optics which is limited in smartphones to keep phone thickness low.
    
    Lensless cameras can be a viable option to overcome size limitations in applications like smartphone photography or tiny robotics. These cameras are generally low-cost, require simpler construction, and are much smaller in length from the sensor as opposed to the multi-lens stack used to direct light and correct for optical distortions.
    In the past few years, research in amplitude \cite{flatcam} or phase mask \cite{diffusercam, antipa2018diffusercam} lensless cameras has gained popularity. This paper revisits the original type of lensless camera, the pinhole, as the more ideal candidate for lensless photography. As shown in Table~\ref{tab:lensless-conceptual}, while the amplitude and phase mask systems are equally low-cost, compact, and have great advantages in light throughput compared to the lensless pinhole camera, they require more precise calibration and reconstruction, and commonly produce lower perceived image quality, making them more suitable for machine vision or scientific imaging applications \cite{flatcamFace}, instead of everyday photography.

    There are additional advantages of pinhole cameras, over their lens-based counterparts, that make them an appealing option for consumer-based photography applications. Pinhole cameras capture an infinite depth-of-field, useful for applications that require all-in-focus images and offer greater freedom from optical distortion due to lenses~\cite{Young:71}. The smaller size advantage of lensless pinhole cameras may also allow implementing larger sensors for smartphone photography, such as micro 4/3, APS-C, or full-frame sensors. These advantages yield a potential direction for pinhole cameras to be used for smartphone photography. 
 
    However, there are several limiting factors to realizing pinhole cameras for conventional smartphones. Pinhole cameras have very low light throughput (eg. f/200) which introduces higher levels of noise. To appropriately expose an image, long shutter speeds are required which can introduce motion-blur artifacts for moving scenes. Further, the sharpness of the captured image is dictated by the level of optical blur caused by the diffraction of light through the pinhole. While both these factors have limited the deployment of practical pinhole cameras in the past, recent advances in image processing technology including higher sensitivity sensors for low-light imaging, better denoising and deblurring algorithms, and data-driven/machine learning methods present an opportunity to revisit the pinhole camera. 
    
    \renewcommand{\arraystretch}{1}
\begin{table}[h!]
    \caption{Comparison of lensless imaging systems}
    \label{tab:lensless-conceptual}
    \centering
    {\small
    \begin{tabular}{p{0.23\linewidth}p{0.20\linewidth}p{0.20\linewidth}p{0.20\linewidth}}
        \hline\hline
        \multicolumn{1}{l}{} & Amp. Mask    & Phase Mask    & Pinhole       \\ 
        \hline
        Rep. Work            & FlatCam      & DiffuserCam   & This work     \\ 
        \hline
        Mask Cost            & Low          & Lower         & Lowest        \\ 
        \hline
        Light Throughput     & Mid          & High          & Low           \\ 
        \hline
        Reconstruction       & Mid          & High          & Low           \\
        Complexity\\
        \hline
        Perceived Quality    & Low          & Low           & High          \\ 
        \hline
        Calibration          & Req'd        & Req'd         & Opt.          \\ 
        \hline
        Suitable App.        & \multicolumn{2}{l}{Machine vision, sci. imaging} & \multicolumn{1}{l}{Photography} \\ 
        \hline\hline
    \end{tabular}
    }
\end{table}

    In this paper, we aim to establish the pinhole camera as a compelling candidate for lensless photography, particularly for size-limited applications. We propose a full system pipeline for improving the quality of pinhole photography at short exposures. We leverage both imaging physics and data-driven networks to build this pipeline. A big obstacle for any deblur (or deconvolution operation in general) is noise which affects pinhole cameras due to their small aperture size. However, the aperture function of the pinhole is a circle, which results in an optical point spread function (PSF) of an \textit{ideal low pass filter (LPF)} of sinc on the imaging plane~\cite{goodman2005introduction}. Thus any signal higher than the cutoff frequency for this LPF must be noise, and we can simply apply a matching LPF in the digital domain to remove any high-frequency noise without affecting the low-frequency signal.

    We leverage this information about a pinhole's optical point spread function (PSF) to design an image restoration pipeline consisting of joint denoising and deblurring for low-light pinhole images. This pipeline is trained on synthetic data but shown to effectively generalize to real-world data, and uses known optical priors including the point spread function (PSF) of the pinhole to help constrain the network's loss functions. We define practical pinhole photography to be images captured at a usable exposure time of 1/30s, while allowing freedom of a higher ISO which can later be denoised through the pipeline. Our specific contributions include the following:
    
    \begin{itemize}
        \item An end-to-end imaging pipeline for practical pinhole photography with joint denoising and deblurring for low-light capture
        \item Reblur losses with a pinhole point spread function (PSF) for improved performance
        \item A real-world pinhole image dataset with measured HDR PSF suitable for generating synthetic data  
        \item Ablation studies for low-exposure images, including the effects of ISO and exposure time.
    \end{itemize}
    
    We validate our pipeline by comparing against both conventional denoising/deblurring algorithms, traditional optimization-based lensless reconstruction algorithms, and end-to-end lensless camera solutions such as diffusion or coded mask cameras~\cite{antipa2018diffusercam,asif2016flatcam}. Our method can handle high-resolution captures of 20 megapixels which is much higher than demonstrated by other lensless camera systems. Our main argument is that practical 2D lensless photography can be achieved with the relatively simpler pinhole camera as opposed to multiplexed imaging methods. 
    We hope this work renews interest in practical pinhole photography in general. 
    \\
\section{Related Work}
\label{sec:related}

    \paragraph{Pinhole cameras.} 
        Pinhole cameras have been around since antiquity with the first recorded mentions by the Chinese philosopher Mozi, Aristotle, Euclid, and Ibn al-Haytham among others~\cite{john1981camera}. During the Renaissance,  Giambattista della Porta, Leonardo da Vinci, and Johannes Kepler can be credited with popularizing the camera obscura~\cite{della1714magia,da1894codex}. However, with the invention of lenses, film, and digital sensors, pinhole cameras largely fell out of common use due to their low-light and optical blur quality. 

        There has been much research on pinhole optics, particularly in characterizing effects such as diffraction on image quality~\cite{Young:71}. The optimum size for the pinhole for mitigating optical blur has been determined for large pinholes using ray optics~\cite{hardy1932principles} and small pinholes using Fraunhofer diffraction~\cite{kingslake}. Much work has gone into examining this tradeoff between ray and diffraction optics for pinhole size~\cite{Sayanagi:67}. There has also been extensive study into the transfer function of pinhole cameras~\cite{Swing:68} and the preservation of spatial frequencies. In this paper, we show that such traditional optical tradeoffs can be overcome with the use of computational imaging techniques. 
    
        Pinhole cameras have been used extensively in scientific imaging applications~\cite{Newman:66,Druart:09,IVANOV1999729}. Computer vision owes a large debt to the pinhole camera, particularly in the camera model that underpins many geometric vision algorithms~\cite{Sturm2014}. However, papers which leverage actual pinhole cameras, as opposed to the model, for computer vision have been sparse in the literature. Some recent inspiring work has included accidental pinhole and pinspeck cameras~\cite{torralba2012accidental,accidentalpinhole} showing that simple image processing could be used for apertures naturally occurring in an environment. 

    \paragraph{Coded Lensless Cameras.} 
        Instead of single pinhole cameras, a bulk of research in the computational imaging and photography community has multi-aperture or multiplexed lensless imaging systems. These systems typically allow more light throughput but typically at the expense of worse reconstructed 2D image quality, as we will demonstrate later in our experimental comparison to some systems. However, these systems do have the express advantage of capturing 3D information through their optical multiplexing which is useful for certain applications. 

        Recently, lensless cameras with an optical element above the sensor have been introduced where these optical elements are thin and scalable to a small size~\cite{tanida2001thin,flatcam,diffusercam,gill2011microscale}. FlatCam~\cite{flatcam} uses a coded amplitude mask to multiplex light from the scene onto the sensor and then reconstructs the image in post-processing. Since amplitude masks lose some light efficiency, newer designs have featured phase masks~\cite{boominathan2020phlatcam,wu2019phasecam3d} for improved performance. Alternatively, diffraction gratings~\cite{gill2011microscale,gill2013lensless,hirsch2014switchable} and Fresnel plates~\cite{tajima2017lensless} have been used to achieve small form factors. Finally, diffusion layers that scatter light onto the sensor have been deployed for lensless imaging systems~\cite{diffusercam}. 
        
        To reconstruct lensless images, optimization algorithms are typically deployed to solve the inverse problem. These typically include a variation on either alternating direction method of multipliers (ADMM)~\cite{boyd2011distributed} or some regularized $\ell_1$~\cite{beck2009fast} or total variation regularization~\cite{rudin1992nonlinear}, which are adapted for lensless imaging~\cite{flatcam,diffusercam}. Recently deep learning has shown superior performance at lensless image reconstruction and other vision tasks~\cite{LenslessLearningSinha,li2018imaging,li2018deep}. We compare against the end-to-end lensless camera systems for FlatNet~\cite{khan2019towards} and DiffuserCam~\cite{Monakhova2019learned} in this paper. 

    \paragraph{Image Deblurring and Denoising.} 
        Image denoising and deblurring are traditional low-level vision tasks with a rich history of research. For denoising, we point the reader to \cite{motwani2004survey} for a thorough survey of traditional algorithms. Block-matching-based approaches such as BM3D~\cite{dabov2007image} and non-local mean~\cite{buades2005non, lebrun2013nonlocal} are top-performing traditional methods widely used nowadays, but are recently exceeded by modern deep learning networks~\cite{zhang2017beyond, zhang2018ffdnet, zhang2017learning}. Leveraging these advances in image denoising research, recent work has changed the photography landscape with successful applications in low-light and burst photography~\cite{chen2018learning, burstphotography}. While many denoisers, including learning-based ones, consider a sensor's noise formation model to improve the results on real-world images~\cite{guo2019toward, abdelhamed2019noise}, they rarely leverage domain knowledge of the optics to denoise better. In contrast, we utilize the optics' frequency response to further improve our learning-based denoiser. While we only focus on pinhole imaging here, this idea can be easily extended to any diffraction-limited system, an increasingly important scenario as pixel size shrinks for modern sensors. 

        For deblurring, it is commonly tackled through either blind or non-blind image deconvolution~\cite{campisi2017blind, kundur1996blind}. For imaging systems, motion blur and defocus blur are the two large categories of deblurring research. Techniques to handle motion blur in photography include traditional~\cite{nayar2004motion,cho2009fast} and neural methods~\cite{chakrabarti2016neural, xu2013unnatural,eboli2020end} including the DeblurGAN architectures~\cite{kupyn2018deblurgan,kupyn2019deblurGANv2} with high-quality performance. Defocus deblurring research has focused primarily on deconvolution problems with applications in microscopy~\cite{sarder2006deconvolution}, dual pixels~\cite{abuolaim2020defocus}, and coded aperture systems~\cite{zhou2009good,zhou2011coded}. In addition, reblur loss for motion blur~\cite{michaeli2014blind, chen2018reblur2deblur} is a technique where the static image is reblurred to match the input data. Reblur losses have also been used for lensless cameras~\cite{Monakhova2019learned,rego2021robust}. In our paper, we also deploy a reblur loss based on our measured optical point spread function. 

        Joint denoise/deblur methods have also been achieved for traditional lensed camera photography, particularly to tackle low-light noise and motion blur through burst capture~\cite{liba2019handheld}. 
        However, we note that there is still relatively less literature on joint denoising and deblurring. This is probably due to the competing natures of the two tasks: one is aimed at reducing high-frequency content (noise) in the image, and the other aims to restore high-frequency content (spatial details) lost due to optical blur. We also observed this phenomenon that denoising and deblurring algorithms can sometimes undo one another's effects, and thus it takes careful tuning and understanding of the frequency content of the signal to make these pipelines work together. That is indeed what we show in this paper by leveraging the frequency cutoff of the optical PSF of pinhole cameras to help deblurring perform better in practice. 
\section{Approach}
\label{sec:approach}

    Our proposed approach involves four critical components: (1) accurate optical point spread function (PSF) capture and modeling, (2) a denoising architecture to improve the low-light captures, (3) high-quality deblurring to restore image details lost due to optical blur, and (4) a reblur loss using the PSF to help jointly train the architecture. A summary of our pipeline can be visualized in Figure~\ref{fig:pinhole-pipeline}. 

    \subsection{Forward Image Model with Optical Point Spread Function}
 
        \begin{figure}
            \centering
            \includegraphics[width=0.45\textwidth]{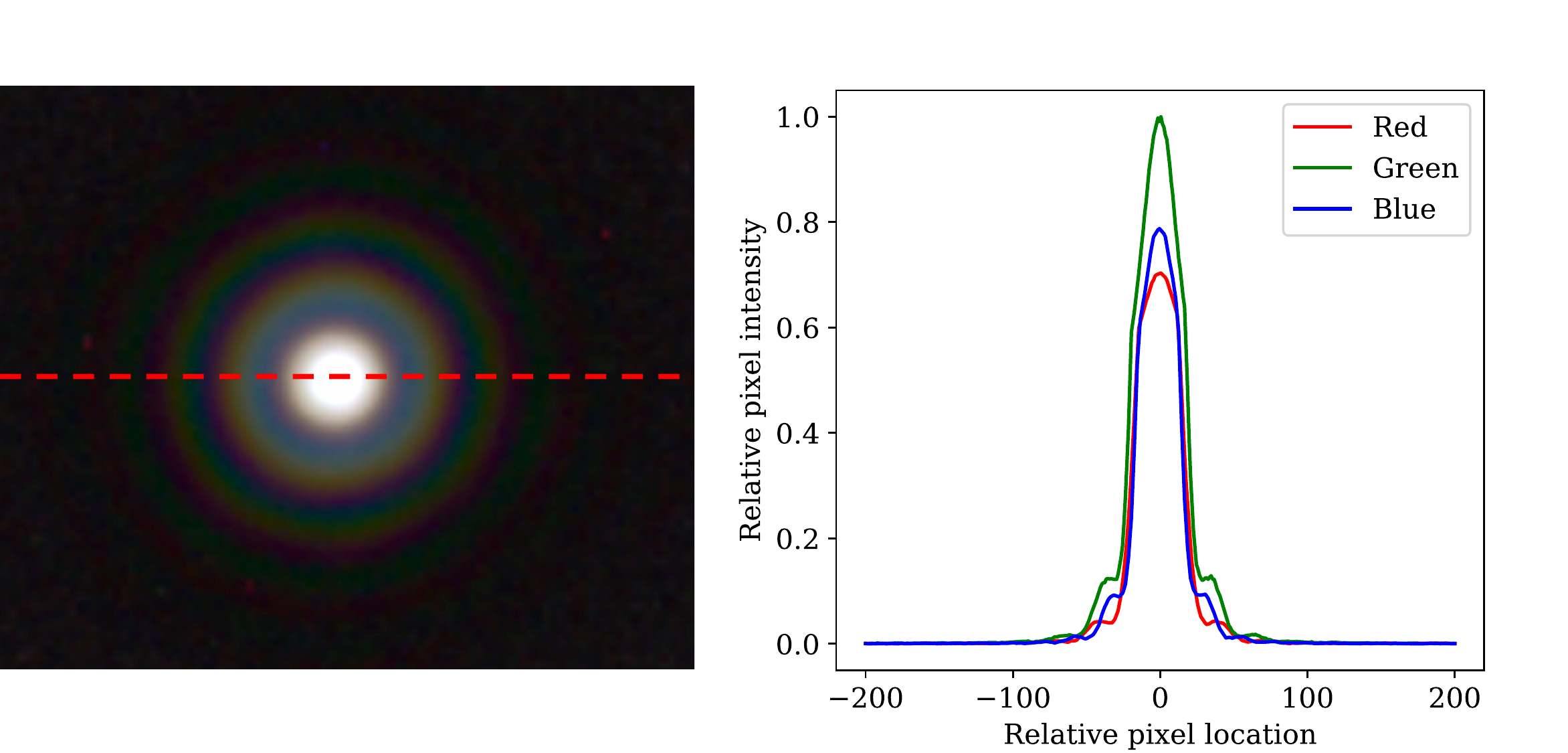}
            \caption{\textbf{System PSF:} The measured PSF of the pinhole is shown. Left is a visualization of the 2D PSF. Due to the HDR nature of the PSF, it is tone-mapped using \cite{debevec1997recovering}. Right is a 1D dissection of the center of the PSF. }
            \label{fig:psf}
        \end{figure}
 
        The optical point spread function (PSF) represents the impulse response of our pinhole imaging system. It determines how a single infinitely small point source in the scene is received at the image sensor. Due to the diffraction effect, the image for a point source is a spatially spread out bright disk surround by concentric rings of light (an Airy pattern). For a wavelength of light $\lambda$, an imaging system with focal length $f$ and aperture of radius $R$ will produce at the sensor plane an Airy disk width $w$ shown by,
        \begin{equation}
            w = 1.22 \frac{\lambda f}{R}.
        \end{equation}
        
        The PSF of our pinhole imaging system (i.e. the intensity distribution of the Airy pattern) can be modeled by the Fraunhofer diffraction of the pinhole aperture. For a circular aperture with radius $R$, the PSF is given by~\cite{goodman2005introduction}:
        \begin{equation}
            P(x,y) = \left( \dfrac{A}{\lambda z} \right)^2 
                    \left[ 2\dfrac{J_{1}(kR\sqrt{x^2+y^2}/z)}{kR\sqrt{x^2+y^2}/z} \right]^2,
        \end{equation}

        where $(x,y)$ is the spatial coordinate on the observation plane, $A = \pi R^2$ is the area of the aperture, $z$ is the distance from the aperture to the observation plane, $J_1$ is a Bessel function of the first order and the first kind, and $k = 2\pi / \lambda$ is the wavenumber where $\lambda$ is the light wavelength. The Fraunhofer approximation is usually only valid for far-field observations, i.e. at a distance $z \gg R^2/\lambda$ which is satisfied by our pinhole camera system. 
              
        The forward model formulates how an image is formed from an object or scene by accounting for the optical response of the camera from the PSF. Specifically, the image is the result of a convolution of the object and the PSF, represented as, 
        \begin{equation}
            i(x,y) = o(x,y) \ast P(x,y),
        \label{eq:forward-model}
        \end{equation}
        where the image $i$, object $o$, and PSF are denoted as functions of position $(x, y)$ in the spatial domain. It's also important to note that, in practice, the image sensor introduces noise that is added to the final image and can be accounted for by adding the noise model to Eq.\ref{eq:forward-model}, i.e. $i = o \ast P + \mathcal{\eta}$. 
        
        This forward model is extensively used in our paper including to help create synthetic data for training our network as well as a reblur loss to check the consistency of our network's image restoration with given camera measurements. 

    \subsection{Denoise Module} 
        Due to the small aperture of a pinhole, the resulting captured images will have a high amount of noise with photon shot noise dominating~\cite{snrreference,jayasuriya2021}. Thus our first step in our pipeline (after demosaicing the RAW image), is to perform denoising. One big advantage to the blur caused by the pinhole is that it creates a global frequency limit to the true image intensity, and thus any signal above the PSF's frequency cutoff is attributable to noise. 
        We observed in practice that even a simple, ideal low-pass filter with the same frequency cutoff used for denoising gave high-quality results when the image was passed to subsequent deblurring. 
        
        However, since noise can also occur at lower and middle frequencies, we utilize a denoise network based on the FFDNet architecture~\cite{zhang2018ffdnet} to further mitigate the noise. Our choice of FFDNet was informed by an ablation study of various neural network architectures which we present in Sec.~\ref{sec:results}. As discussed in \cite{tassano2019analysis}, FFDNet's unique and simple initial pixel-shuffling down-sample layer doubles the receptive field without increasing the network size, resulting in faster execution time and smaller memory footprint, while being able to handle various noise types. To train this architecture, we utilize synthetic data and simulate photon and Gaussian read noise for various ISOs for the camera as detailed in Sec.~\ref{sec:implementation}.

    \subsection{Deblur Module}
        Once the denoised image is obtained from the denoise module, we utilize the state-of-the-art DeblurGANv2~\cite{kupyn2019deblurGANv2} to restore the blurry pinhole image, given its superior performance in our ablation study in Sec.~\ref{sec:results}.
        Since the original model is trained primarily for motion blur, the pre-trained model is not suitable for the optical blur in pinhole images. For our pipeline, we train this network on our own synthesized pinhole datasets for it to perform deblurring on pinhole images.

    \subsection{Reblur Loss}
        One key challenge to a joint denoising and deblurring architecture is that deblurring can undo the effects of denoising and reintroduce noise-like artifacts into the final image. This is particularly the case in background patches in the image, where the deblur module needs to infer the missing data without many textures or patterns to infer from. 
        
        To alleviate this, we introduce joint training of our network using a \textit{reblur loss}. Reblur losses have been introduced in other contexts including motion blur~\cite{chen2018reblur2deblur, michaeli2014blind} and lensless imaging~\cite{Monakhova2019learned,rego2021robust}. We utilize the captured PSF from our camera system and perform the convolution of our generated output from the network to form an estimated reblurred image $\Hat{i}$. Then we utilize an MSE loss between $\Hat{i}$ and the original lensless image $i$ to fine-tune our network during training. Thanks to the simplicity of the pinhole imaging model, the PSF convolution reblur model is highly accurate and can be measured easily, making the reblur loss more effective than motion blur cases \cite{michaeli2014blind, chen2018reblur2deblur}.
        
\section{Implementation}
\label{sec:implementation}
    
    \subsection{Training Dataset}
        To train our network architecture, we use the HDR+ subset~\cite{burstphotography} containing, $153$ images of $4048\times3036$ resolution. These are trained with $256\times256$ size patches. We create a simulated blurred version of this dataset as our input during training by convolving the real captured PSF with the original data. The original data is then used as ground truth while training the network. We implement both photon shot noise and Gaussian read noise in our simulator, using a realistic noise simulator with parameters set for various ISOs which we randomly toggle during training. 
    
        \begin{figure*}[ht]
            \begin{center}
                \includegraphics[width=0.9\linewidth]{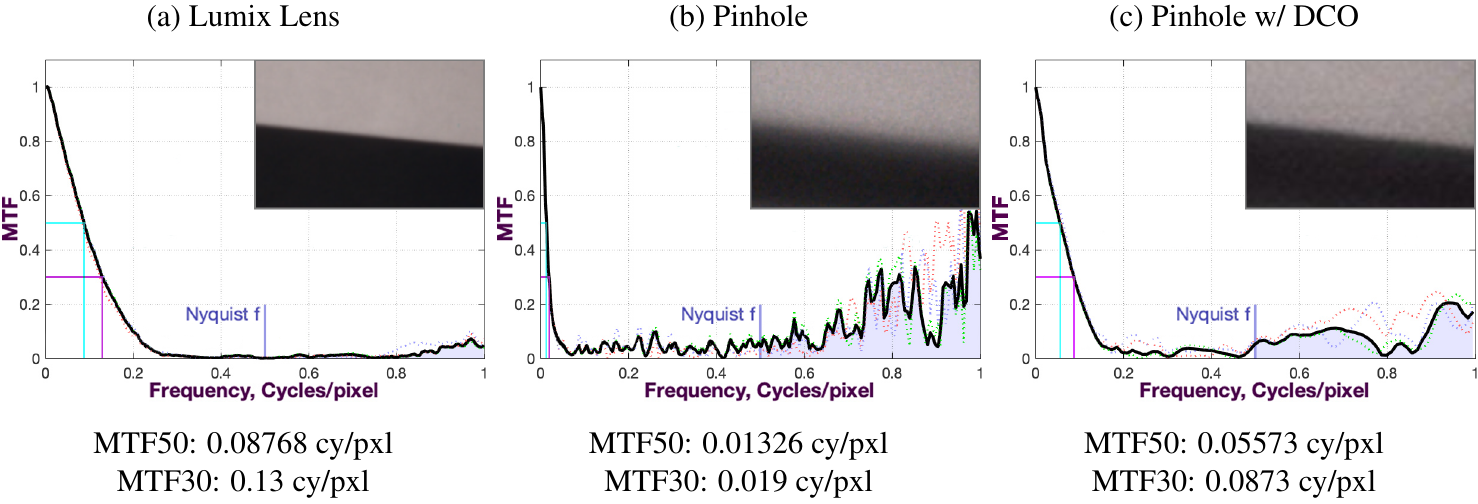}
            \end{center}
            \vspace{-3mm}
            \caption{MTFs of a Panasonic Lens, Thingyfy Pinhole, and Pinhole w/ DCO, obtained with Imatest on the ISO12233 chart.}
            \vspace{-3mm}
            \label{fig:MTF}
        \end{figure*}
        
    \subsection{Implementation details.} 
        Our training procedure consisted of the following:
        (1) separately training the denoise module on blurred images with noise, and the deblur module with the subsequent denoised output images. Both the FFDNet and DeblurGANv2 are initially trained from scratch on the synthetic dataset.
        (2) After separate training, joint finetuning of the architecture is done while incorporating the reblur loss. 
        
        Our system was trained on two NVIDIA 2080Ti GPUs. While training, we use a batch size of $1$ image which is cropped to multiple patches of size $256\times256$ per iteration. For both, the generator and discriminator, we use the Adam~\cite{adam} optimizer using a learning rate of $1e-4$ for 50 epochs then reduce the learning rate with a linear decay from $1e-4$ to $1e-7$  epochs for another 200 epochs. For the generator loss, we assign weights to $\mathcal{L}_{MSE}$, $\mathcal{L}_{perc.}$, and $\mathcal{L}_{adv}$ of $\lambda_{MSE}=0.5$, $\lambda_{perc.}=0.006$, and $\lambda_{adv}=1e-3$ which is used for both the generator and discriminator adversarial losses.

    \subsection{Real data capture} 
        Additionally, we capture some real pinhole camera images which we use for both quantitative and qualitative analysis of our pipeline. These are captured using a Panasonic Lumix G85 mirrorless camera with a micro 4/3 sensor and the Thingyfy Pinhole Pro using a pinhole diameter of $0.15$mm. The images are captured at 1/30 second exposure time with $3200$ ISO. From the RAW image, we only apply black level correction and demosaicing to keep the tone-mapping linear. 
        
        For our real data capture, it is not trivial to capture ground truth images. Even using a beamsplitter to optically align a camera with a lens with the pinhole camera does not suffice, as the lens image will have optical aberrations and a finite depth-of-field. Thus in this paper, we choose to train on synthetic data and only use our network at inference time on real data. This has the distinct advantage of showing the generalizability of our network. For real data, due to a possible mismatch between real noise and our noise simulations, we lowpass filter the images with the PSF before running through our denoise and deblur modules. We found this helps improve results for our network as the denoising network is expecting noise statistics similar to that of the synthetic data. 
        
        To compare against other lensless camera systems, namely the FlatCam and DiffuserCam, we also capture scenes displayed from a monitor similar to these papers. We display their test dataset and capture these with our pinhole camera. Note that our network is not trained on their datasets, but only tested on their images at inference.

\section{Results}
\label{sec:results}
    
    \subsection{End-to-end results}
        \paragraph{Tradeoff between pinhole and lenses.}
            Smartphone photography scaling requires larger sensors, forcing lens size to scale exponentially and increase their size/weight. Lenses also have inherent aberrations (spherical, chromatic, etc.), while lensless alternatives are much more lightweight and free of those aberrations. Thus, we do not compare to lens systems, but rather, other lensless work instead. However, we did conduct a modulation transfer function (MTF) analysis shown in Figure~\ref{fig:MTF}. Our DCO improves the resolution in terms of MTF50 of the pinhole system by 4.2x, achieving 63.5\% of the lens counterpart.
           
           \begin{figure*}[t!]
                \centering
                \includegraphics[width=\textwidth]{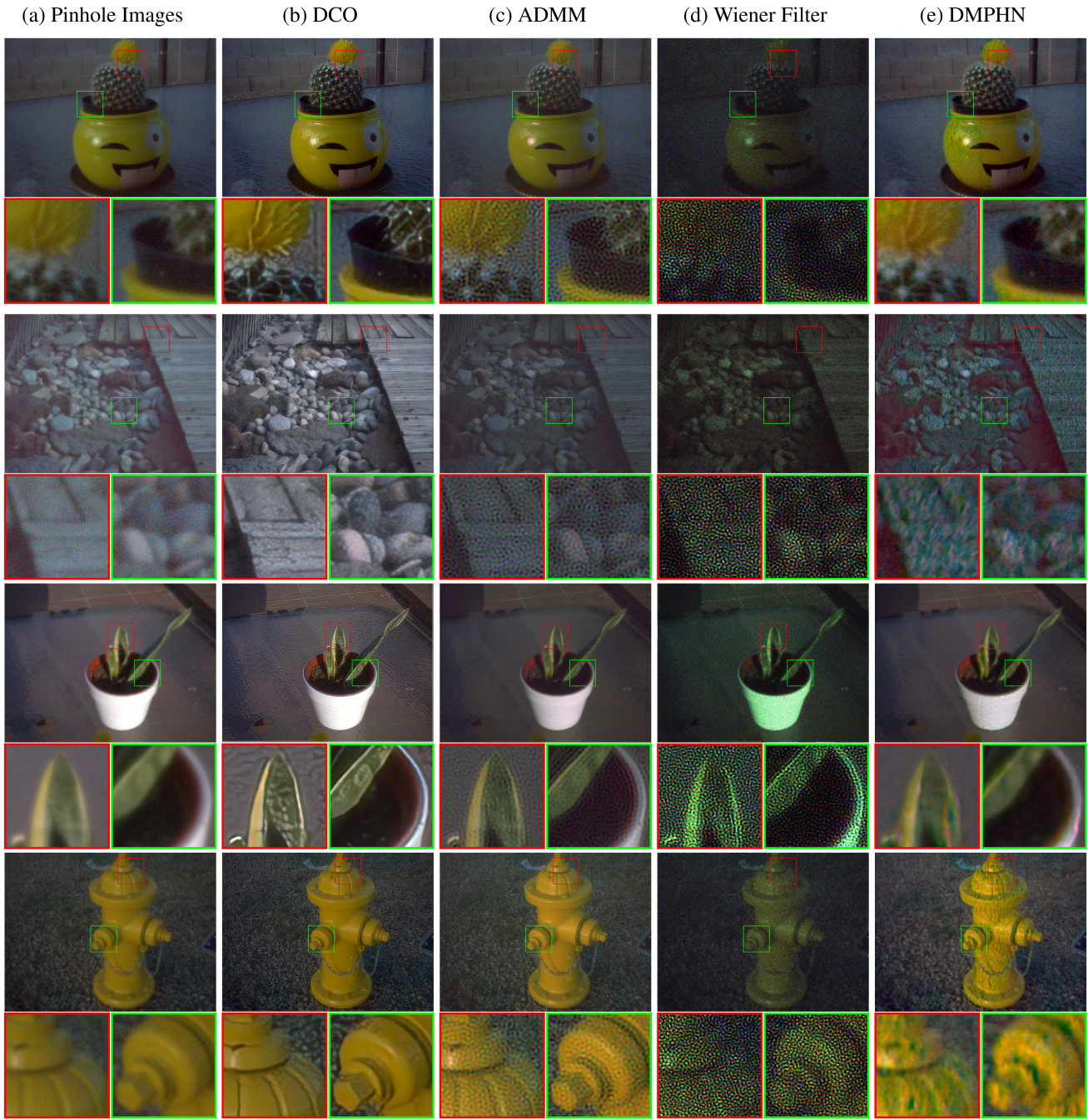}
                \caption{Real captured results for pinhole camera and our DCO results. Note the spatial frequencies recovered using our network, which has not been trained on real data, but transferred from a synthetic dataset.}
                \label{fig:pinhole-real}
            \end{figure*} 
            
        \paragraph{Real-world scene results.}
            We show results on real-world scenes captured by pinhole camera and restored by our proposed pipeline in Figure~\ref{fig:palm-result}. Note that our network has not been trained on real-world data, but solely on synthetic data simulated from the HDR dataset. As you can see in (c) and (d), the denoising and deblurring networks are respectively reducing the noise in the images and sharpening the resulting image. The resulting image in (d) is comparatively sharp as compared to a lens-camera result in (e). 
            
            We present some results from real pinhole captured low-light images in Figure~\ref{fig:pinhole-real}, using the Panasonic m4/3 camera and Thingyfy Pinhole Pro. 
            These feature a variety of scenes with different objects and textures. Note that these images have no corresponding ground truth as we are not imaging a monitor like our lensless system comparisons. We see our captured pinhole images in (a), the resulting DCO reconstructions in (b), and comparisons to two off-the-shelf optimization methods, ADMM (c) and a simple Wiener filter (d) for deconvolution/deblurring, and (e) DMPHN~\cite{zhang2019deep}, a modern deblurring neural network. Note how our method recovers back sharp, high-frequency details with less noise or artifacts as compared to ADMM, Wiener filter, and DMPHN. All these images were taken at 1/30 seconds exposure with ISO 3200.
            
            \begin{figure*}[ht!]
                \centering
                \includegraphics[width=0.93\textwidth]{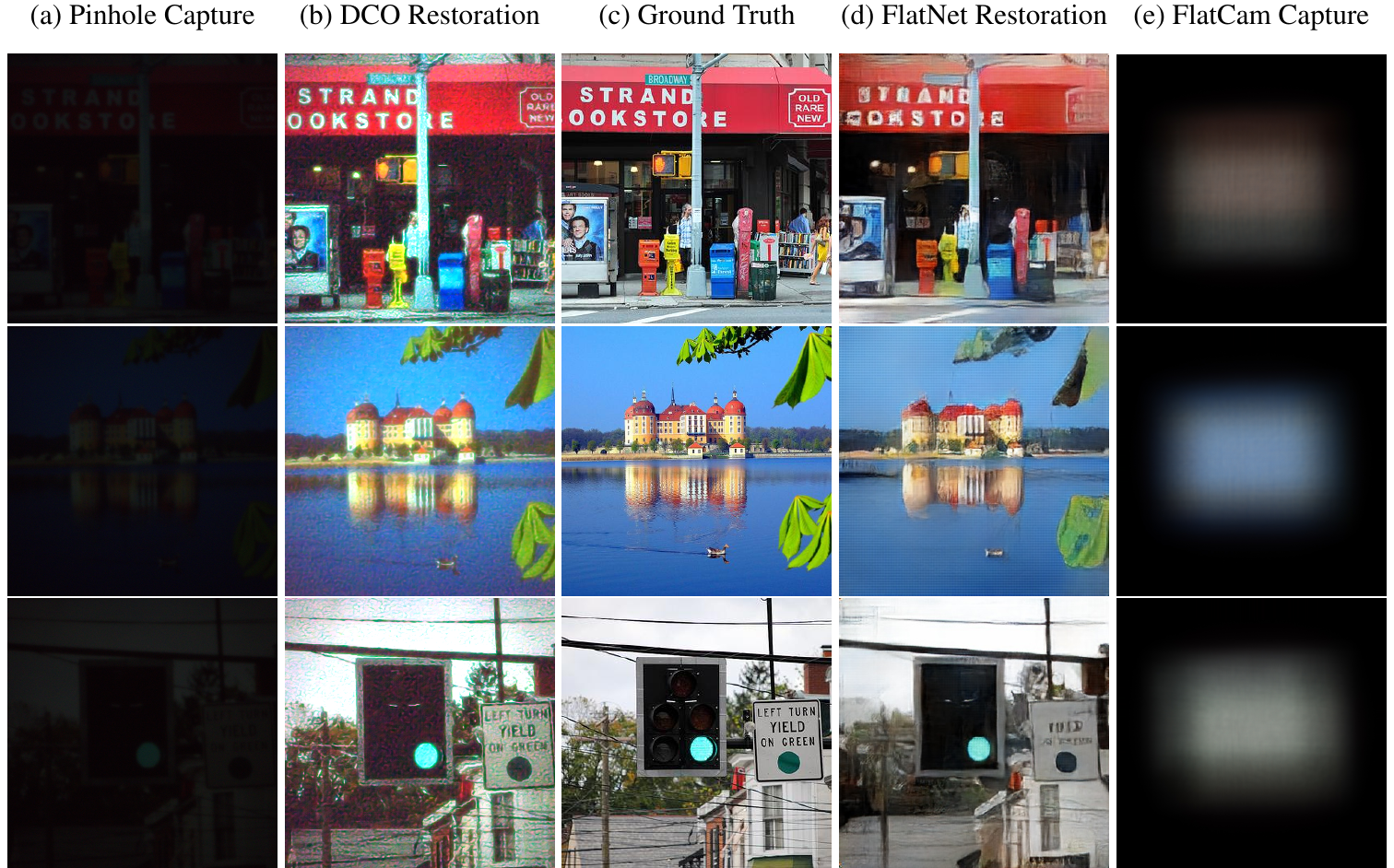}
                \caption{\textbf{Lensless comparisons with FlatNet.} We compare the input and reconstruction of our proposed system (DCO) with FlatCam/FlatNet~\cite{flatcam,Khan2020FlatNet} on monitor captures of the ILSVRC 2012 dataset~\cite{ILSVRC2015}. Note that our input and results are of higher resolution ($1920 \times 1080$) and have higher frequency features without much reconstruction artifacts compared to the FlatNet results from \cite{Khan2020FlatNet}.}
                \label{fig:pinhole-test}
            \end{figure*}

        \paragraph{Display captured results.} 
            Here we perform a comparison with a state-of-art, end-to-end lensless camera system, FlatNet~\cite{Khan2020FlatNet}. We use the testing dataset from~\cite{Khan2020FlatNet} and display the images on a monitor, and then capture it with exposure time 1/30s at ISO 3200 with our hardware pinhole camera. In Figure~\ref{fig:pinhole-test}, we show the qualitative results of this comparison. Note that the pinhole naturally captures less light than FlatCam due to its single aperture. This results in a higher noise level for our DCO reconstruction in (b). However, note that the details of the image are preserved at high spatial frequencies as compared to FlatNet. In FlatNet's reconstructions (d), certain artifacts from the machine learning include warping along the castle's reflection, distorted signs and text, and non-physically realistic deformations.
            
\begin{figure*}
                \centering
                \includegraphics[width=0.94\textwidth]{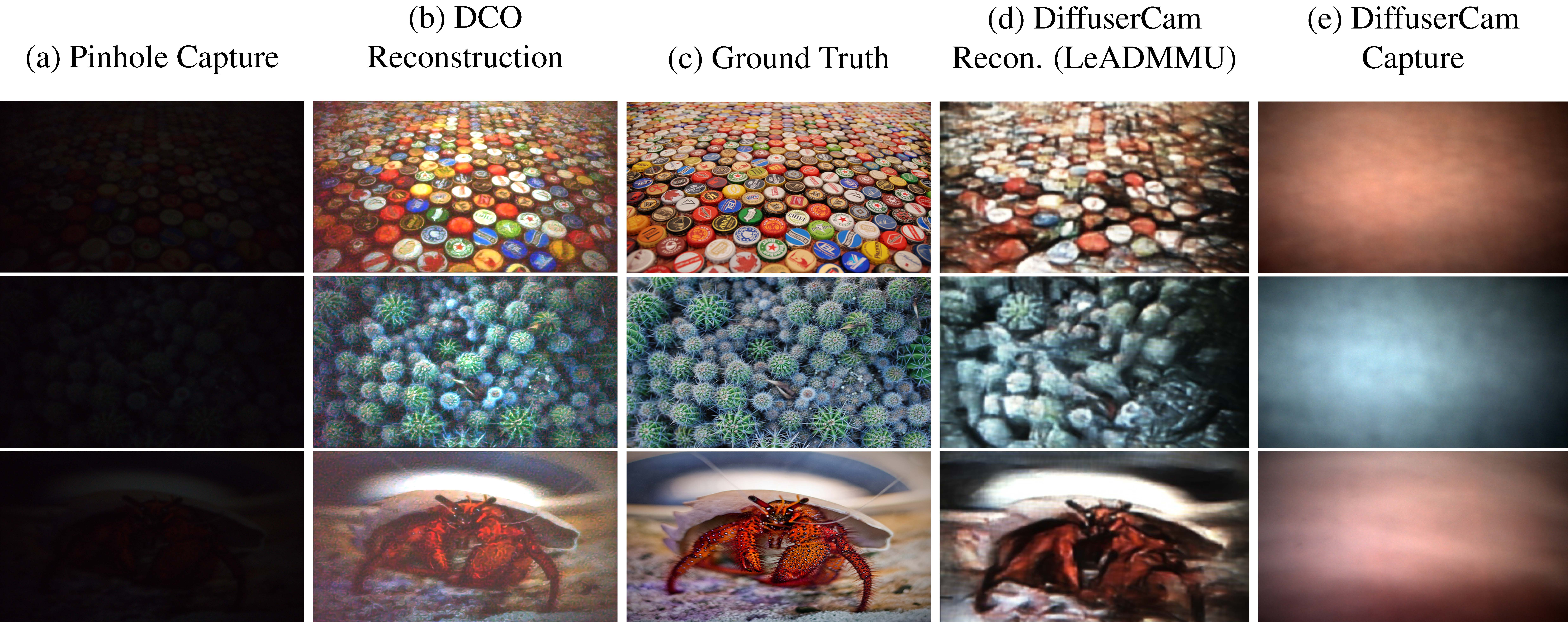}
                \caption{\textbf{Lensless comparison with DiffuserCam.} We compare the input and reconstruction of our proposed system (DCO) with DiffuserCam, a phase mask-based lensless imaging system, on the DiffuserCam datasets. Note that though our input and results are of higher resolution ($1920 \times 1080$) compared to the DiffuserCam results from \cite{Monakhova2019learned}.}
                \label{fig:diffcam-results}
            \end{figure*}

            In Figure \ref{fig:diffcam-results}, we compare the results from our proposed lensless system (DCO) with a phase mask-based lensless imaging system, DiffuserCam~\cite{Monakhova2019learned}. Similar to DiffuserCam, we display the ground-truth images on an LCD monitor, capture the images using the lensless pinhole camera, and process them using the DCO framework. Note that DiffuserCam's raw images are captured at $1920 \times 1080$, reconstructed at $480\times270$, and cropped to $380\times 210$. For their evaluation, the reference images are captured using a lensed camera through a beam splitter, then resized and cropped to the same $480 \times 270$ size.  In contrast, our raw captures and reconstructions are all conducted at $1920 \times 1080$, the reference images in our evaluation are the original displayed images, resized to $1920 \times 1080$. Our results show that pinhole cameras are a viable alternative to coded aperture cameras for traditional 2D photography applications.
            
            In Table~\ref{tab:lensless}, we report our PSNR and SSIM numbers as compared to FlatNet and DiffuserCam as reported by those papers~\cite{Khan2020FlatNet,Monakhova2019learned}. We score higher in PSNR than both methods by a small margin which shows the advantage of our opinion.
            Note that an SSIM metric evaluation was not available in DiffuserCam's paper.
            
            \renewcommand{\arraystretch}{1.5}
\begin{table}[h!]
    \caption{Quantitative comparison of lensless systems.}
    \label{tab:model-compare}
    \centering
    \begin{tabular}{cccc}
        \hline\hline
                    & FlatCam   & DiffuserCam   & DCO/Pinhole \\
        \hline
        PSNR (dB)   & 19.62     & 21.31         &  22.31     \\
        SSIM        & 0.64      & \textit{N/A}  &  0.634   \\
        \hline\hline
    \end{tabular}
    \vspace{-4mm}
    \label{tab:lensless}
\end{table}

        \paragraph{Video Reconstruction.}
            Finally, we show a real-world video captured from a monitor at 1/30 seconds exposure with our pinhole camera. Note that since our current hardware cannot support video capture of RAW frames, we captured each frame individually by using a synchronized shutter. 
            As one can see from the video frames, the input pinhole images are blurry and suffer from noise effects. Our method improves sharpness around object edges and enhanced textures in the wood, without noticeable noise degradation.
            
            We do note some limitations to applying our network to video frames. The resulting video suffers from some temporal artifacts including flickering. While there is still room to improve upon video results, we still feel this demonstrates the improvement benefits of our DCO network in practice.

            \begin{figure}[ht!]
                \begin{center}
                   \includegraphics[width=\linewidth]{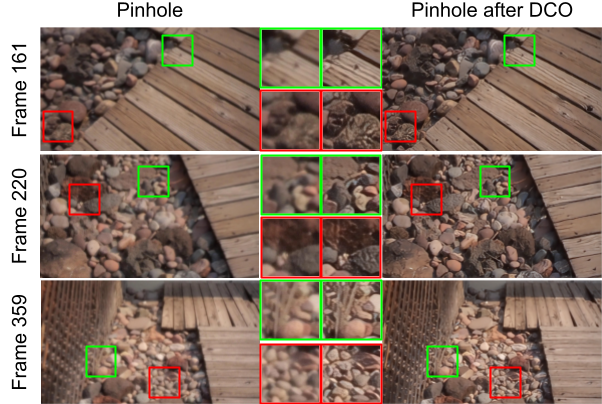}
                \end{center}
                \caption{Frames from a pinhole video capture before and after DCO.}
                \label{fig:video}
            \end{figure}
        
    \subsection{Ablation studies}
        \paragraph{Denoise module.}
            Here we benchmark three denoise network candidates for denoising performance on the synthetic pinhole noise data. Overall, FFDNet performs better than FOCNet, and is on par with FC-AIDE quantitatively but with improved perceptual quality. Therefore, we chose FFDNet for our final pipeline based on this ablation study.

            \renewcommand{\arraystretch}{1.5}
\begin{table}[h!]
    \caption{Quantitative performance comparison of denoise network candidates.}
    \label{tab:denoise-compare}
    \centering
    \begin{tabular}{ccccc}
        \hline\hline
                    & FOCNet    & FC-AIDE   & FFDNet    \\
        \hline
        PSNR (dB)   & 29.2964   & 30.0940   & 30.0915     \\
        SSIM        & 0.7919    & 0.9448    & 0.9452    \\
        \hline\hline
    \end{tabular}
    \vspace{-4mm}
\end{table}

            \begin{figure*}[ht!]
                \centering
                        \includegraphics[width=\textwidth]{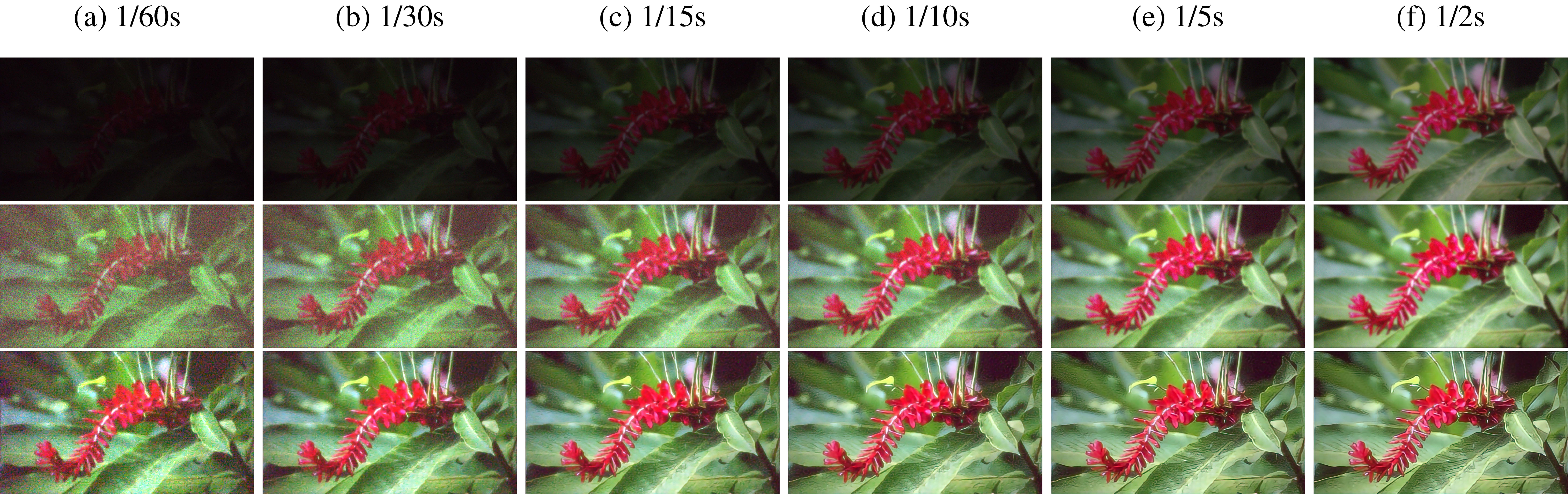}
                \caption{\textbf{Exposure sweep experiment.} We captured data with the ISO fixed as 1600 and perform the reconstruction with the proposed DCO framework. Top row is input, middle row is brightened input, and the last row is DCO reconstruction. The results indicate that light throughput does have a noticeable impact on the final reconstruction quality.}
                \label{fig:exposure-sweep}
                \vspace{-2mm}
            \end{figure*}
            
            \begin{figure*}
                \centering
                \includegraphics[width=\textwidth]{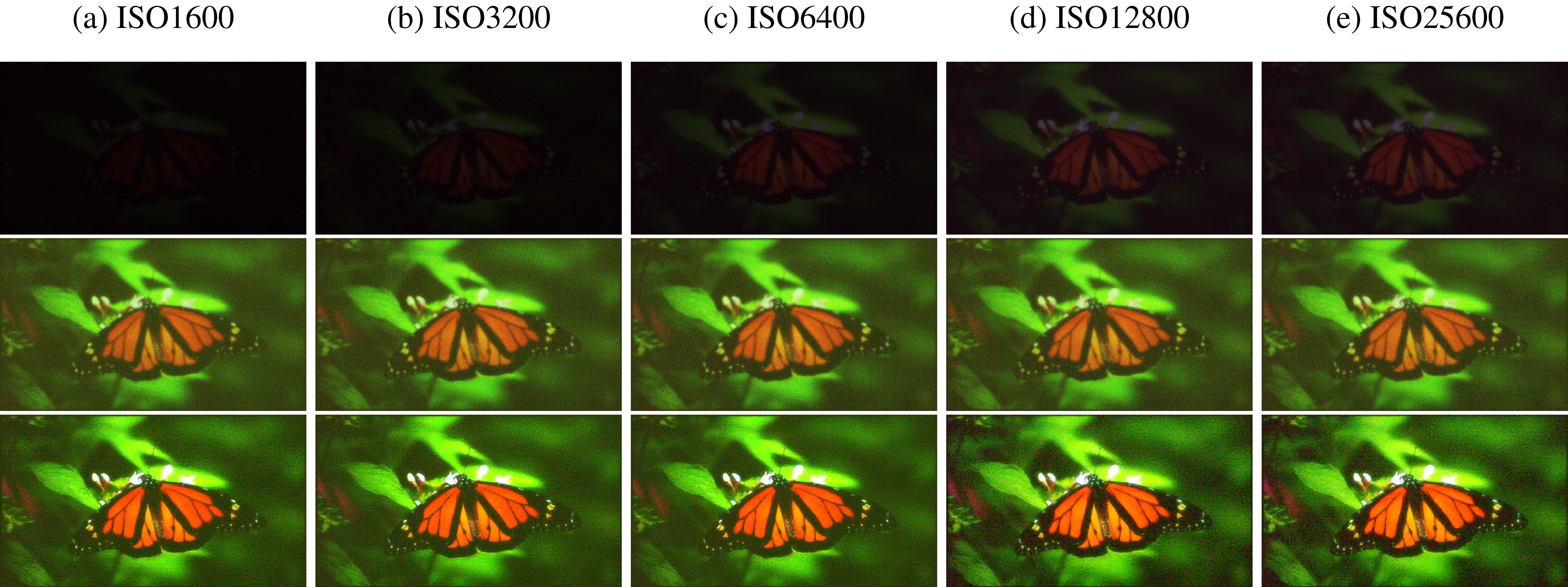}
                \caption{\textbf{ISO sweep experiment.} We captured data with the exposure time fixed as 1/30 seconds and perform the reconstruction with the proposed DCO framework. Top row is input, middle row is brightened input, and the last row is DCO reconstruction. The results indicate that the ISO setting does not impact the final reconstruction results significantly. }
                \label{fig:iso-sweep}
                \vspace{-4mm}
            \end{figure*}
            
        \paragraph{Deblur module.}
            We also performed an ablation study to compare three different deblurring networks for our pipeline on the synthetic pinhole blur data. These networks were DMPHN~\cite{zhang2019deep}, SIUN~\cite{ye2020scale}, and DeblurGANv2~\cite{kupyn2019deblurGANv2}. As we can see from the table, DeblurGANv2 achieved the highest scores and thus was our choice for the deblurring module in our pipeline. 
            
            \renewcommand{\arraystretch}{1.5}
\begin{table}[h!]
    \caption{Quantitative performance comparison of deblur network candidates.}
    \label{tab:deblur-compare}
    \centering
    \begin{tabular}{ccccc}
        \hline\hline
                      & DMPHN    & SIUN        & DeblurGANv2     \\
        \hline
        PSNR (dB)    & 27.273    & 26.234    & 27.385       \\
        SSIM         & 0.6805    &  0.6372    & 0.6827       \\
        \hline\hline
    \end{tabular}
    \vspace{-4mm}
\end{table}

        \paragraph{Exposure time.}
            At the very low light throughput condition the pinhole camera operates in, the noise is dominantly photon noise. To understand photon noise's impact on the reconstruction, we fixed the digital gain setting of the camera to be ISO 6400 and conduct reconstruction at various exposure times. The results are shown in Figure \ref{fig:exposure-sweep}. We do observe that reconstruction, particularly denoising, gets more challenging when light throughput decreases.
        
        \paragraph{ISO.}
            In contrast to the exposure sweep experiment, we also fixed the exposure time to be 1/30 seconds and sweep across different ISO settings from 1600 to 25600. In this case, the light throughput is constant, and read noise is amplified variously depending on the ISO settings. As Figure~\ref{fig:iso-sweep} shows, we found ISO settings do not impact the final reconstruction results noticeably.
\section{Discussion}
\label{sec:discussion}
    
    In this paper, we presented a practical pinhole photography pipeline to restore low-light images. To correct the pinhole camera issues of blur and high-noise from low-light imaging, the pipeline performs denoising using a low-pass filter with a frequency limit prior based on the optical point spread function and then passing through a deep learning-based denoising network FFDNet~\cite{zhang2018ffdnet} and deblurring using a retrained DeblurGANv2~\cite{kupyn2019deblurGANv2} module on our synthesized pinhole dataset. We verify our pipeline on test data from the synthesized HDR+ dataset as well as real-world pinhole captured images to show qualitative improvements for both denoise and deblur problems. We also note that there is still room for improvements to the deblurring results, particularly for high-frequency features.
    
    We plan to extend the work in this paper to include a modified video-specific pipeline as well as options for a learnable low-pass filter. There also may be further avenues for applications to work on past practical pinhole photography for 2D images. Since the exposure of the image can be fixed, HDR imaging for pinhole cameras can be an interesting direction that can further its potential for smartphone photography. Additionally, stereo pinhole photography can open up use cases for 3D depth estimation for SLAM algorithms and robotics in general.

{\small
\bibliographystyle{ieee_fullname}
\bibliography{egbib}
}

\end{document}